\documentclass[12pt,a4paper]{article}
\usepackage[numbers,sort&compress]{natbib}

\usepackage[T1]{fontenc}
\usepackage{mathptmx} % Times-like font
\usepackage{helvet}   % Helvetica-like font

\usepackage{multirow}
\usepackage{algorithm2e} % Add algorithm environment support
\usepackage{listings} % Add code listing support

\usepackage[final]{microtype}

\usepackage[final]{nips_2018}

\usepackage{float}

\usepackage[utf8]{inputenc} % allow utf-8 input
\usepackage{hyperref}       % hyperlinks
\usepackage{url}            % simple URL typesetting
\usepackage{booktabs}       % professional-quality tables
\usepackage{amsfonts}       % blackboard math symbols
\usepackage{nicefrac}       % compact symbols for 1/2, etc.
\usepackage{graphicx}
\usepackage{tabularx}
\usepackage{hyperref}
\usepackage{amsmath}        % for math formulas

\usepackage{fancyhdr}
\usepackage{titlesec}
\titleformat{\section}{\Large\bfseries}{\thesection}{1em}{}
\titlespacing*{\section}{0pt}{3.5ex plus 1ex minus .2ex}{2.3ex plus .2ex}

\title{Refine Thought: A Test-Time Inference Method for Embedding Model Reasoning}

\author{
  \textbf{Guangzhi Wang} \quad \textbf{Kai Li} \quad \textbf{Yinghao Jiao} \quad \textbf{Zhi Liu} \\[0.5em]
  CareerInternational Research Team \\[0.5em]
}

\begin{document}
% \nipsfinalcopy is no longer used
\maketitle

\begin{abstract}

We propose RT (Refine Thought), a method that can enhance the semantic reasoning ability of text embedding models. The method obtains the final semantic representation by running multiple forward passes of the text embedding model. Experiments show that RT achieves significant improvements on semantic reasoning tasks in BRIGHT \cite{su2024bright} and the person job matching benchmark PJBenchmark \footnote{PJBenchmark (Person-Job Benchmark) is a test set for recruitment matching scenarios.} , while maintaining consistent performance on general-purpose semantic understanding tasks such as C-MTEB. Our results indicate that RT is effective because it further activates the semantic reasoning ability learned during pretraining by decoder-only text embedding models(e.g., Qwen3-Embedding-8B \cite{zhang2025qwen3embedding}). RT can be seen as a test-time \cite{snell2024scaling} inference method.

\end{abstract}

\section{Introduction}
\label{sec:introduction}

Neural networks can be seen as both representation learners that automatically acquire hierarchical features and differentiable universal computers \cite{lecun2015deep}. Reasoning is essentially a multi-step computation involving sequential decisions. It includes arithmetic, symbolic, semantic, and spatial reasoning \cite{huang2023reasoning}. Among these, semantic reasoning relies both on multi-step computation and on hierarchical language representations. Most text embedding models perform self-supervised learning on large-scale text via pretraining objectives, which significantly improves hierarchical language representations.Although this greatly boosts performance on general semantic tasks, limitations remain on semantic reasoning tasks that require multi-step computation \cite{su2024bright}.

In text generation, semantic reasoning has been extensively studied. There are two ways to improve reasoning through training: one is structural unrolling, which deepens the network to allow more layers of functional composition and thereby enhances its expressive power \cite{raffel2020t5}; the other is temporal unrolling, which does not increase the depth but instead reuses the same structure across multiple time steps, thereby allowing the model to progressively refine its internal computation \cite{wei2022chain,saunshi2025looped}.

Most text embedding models \cite{xiao2023cpack} obtain semantic representations by encoding the query and the document each with a single forward pass, and then use nearest-neighbor similarity for retrieval or matching. However, when a query requires multi-step computation---such as multiple constraints, cross-sentence integration, or sequential reasoning---a single forward pass fails to capture intermediate reasoning representations and lacks sufficient computational depth.

Research on semantic reasoning within text embedding models is still limited. With the growing demand for RAG \cite{fan2024ragllm} and Agent systems \cite{plaat2025agentic}, LLMs increasingly need to retrieve and reason over internal information. As a result, enhancing the semantic reasoning ability of text embedding models has become a key research challenge. To address this limitation, we propose RT for text embedding models. RT executes multiple forward passes on the query side to obtain semantic reasoning at test time, without modifying model parameters. Across BRIGHT, C-MTEB, and PJBenchmark, our experiments show that RT significantly improves semantic reasoning tasks while remaining stable on general-purpose semantic understanding tasks.

Our contributions are summarized as follows:

\begin{enumerate}
\item We propose RT, a method that improves semantic reasoning in text embedding models without extra training, and provide algorithmic details.

\item We systematically analyze the relationship between the performance gain and task complexity. The results show that the more complex the task, the more pronounced the gains from RT, while performance on simple tasks remains largely unchanged.

\item We validated RT across multiple benchmarks, indicating strong potential for real-world applications.
\end{enumerate}

\section{Related Work}
\label{sec:related-work}

Neural Turing Machine research shows that when models fail to reason, the problem is usually not logical but computational — their computation graphs are not unrolled long enough in time for information to pass through. Complex reasoning requires recursive, variable-length computation over multiple time steps \cite{graves2014ntm}. Chain-of-Thought (CoT) increases text output length, enabling more time steps for multi-step computation \cite{wei2022chain}. DeepSeek-R1 improves reasoning by training models via reinforcement learning to force chain-of-thought outputs \cite{deepseek2025r1}.

Recent theoretical work provides a formal basis for the statement that “temporal unrolling improves arithmetic and symbolic reasoning” \cite{li2024chainofthought}. With fixed network depth and limited precision, allowing a model to generate and exploit a CoT sequence of length $T$ during inference effectively trades time steps for computational depth. This implies that reasoning ability can be systematically improved by extending time steps without altering the network structure.

\section{Method}
\label{sec:method}

\subsection{Problem Definition}

Given a query $x_q$ a set of candidate documents $C={x_d}$, text encoders usually compute embeddings $h_q(x_q)$ and $h_d(x_d)$ through a single forward pass, and then rank documents according to their similarity:

$$\hat{d}=\arg\max_{d\in C}\;\mathrm{sim}\!\big(h_q(x_q),\,h_d(x_d)\big)$$

This approach is computationally efficient, however, when the query $x_q$ involves multiple constraints, cross-sentence information integration, or sequential reasoning, a single forward pass often fails to preserve the underlying reasoning trajectory. We refer to such cases as semantic reasoning tasks. The reasoning process for $x_q$ can be formalized as:

$$
h_t = 
\begin{cases} 
f(x) & \text{if } t = 1 \\[1.5ex]
f\!\left(x,\, h_1, \ldots, h_{t-1}\right) & \text{if } t \in \{2,\dots,T\}
\end{cases}
$$

In this process, the model passes through multiple states to obtain the final semantic representation.

\subsection{The Refine Thought (RT) Method}

We use a decoder-only text embedding model that first performs a forward pass for the query $x_q$. The representations of all tokens before the [EOS] token are mean-pooled to obtain the intermediate representation $h_{t-1}$ for the $t$-th step. Based on this, the process is iteratively repeated for $T$ steps, and the intermediate representations $\{h_1, h_2, \ldots, h_{T-1}\}$ collectively form the reasoning trajectory. The detailed procedure is as follows:

\begin{lstlisting}[language=Python, caption={RT (Refine Thought) Method – Python Pseudocode}, basicstyle=\tiny\ttfamily]
def refine_thought(x_q, T, encoder):
  # Inputs: query x_q, iteration steps T
  # Output: final query embedding h_T

  # Step 0: initial representation (average-pool all tokens before [EOS])
  h_0 = encoder(x_q) 

  # Store intermediate states
  states = [h_0] 

  # Iteratively refine
  for t in range(1, T+1):
      # Step 1: Read from previous state
      prev_state = states[-1]
      
      # Step 2: Update query representation based on all past states
      x_q_with_all_states = concatenate(x_q, states) 
      h_t = encoder(x_q_with_all_states) 
      
      # Step 3: Append for the next round
      states.append(h_t)

  # Return the final state h_T for similarity scoring
  return states[-1] 
\end{lstlisting}

\begin{figure}[htbp]
    \centering
    \includegraphics[width=0.4\textwidth]{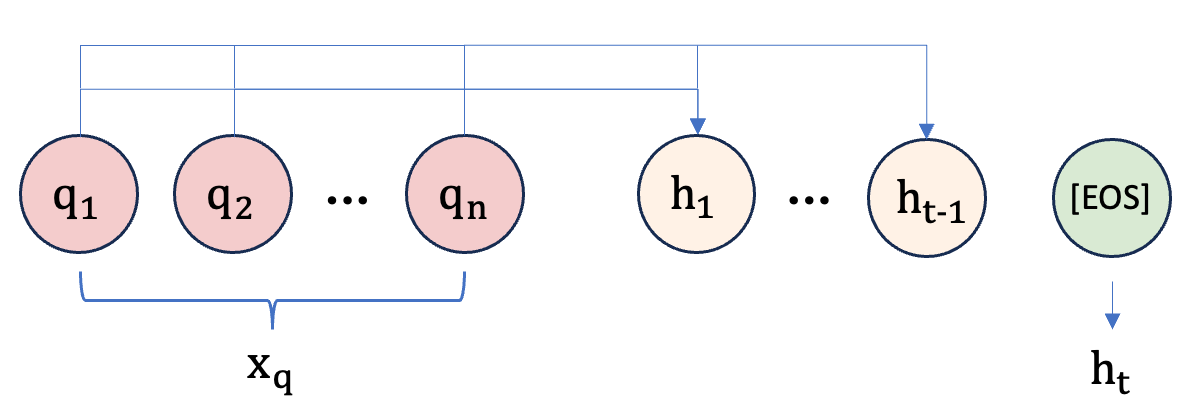}
    \caption{RT(Refine Thought)method overview}
    \label{fig:rt-architecture}
\end{figure}

This process effectively adds extra computation steps during inference, improving the model’s reasoning depth. In contrast to the explicit CoT approach \cite{wei2022chain}, RT performs these “time steps” implicitly in the hidden space \cite{zhu2025latentreasoning}, without generating additional text or altering the document representations or indexes. The schematic of the RT method is shown in Figure~\ref{fig:rt-architecture}.

The step count $T$ is set based on task complexity, with more complex tasks requiring larger $T$.

\section{Experiments}
\label{sec:experiments}

\subsection{Experimental Setup}

We evaluate RT on BRIGHT, C-MTEB, and PJBenchmark, and choose Qwen3-Embedding-8B \cite{zhang2025qwen3embedding} and bge-large-zh-v1.5 \cite{xiao2023cpack} to represent decoder-only and encoder-only text embedding architectures, respectively.

We evaluate three types of tasks:
(i) Semantic Textual Similarity (STS) tasks from C-MTEB, used to verify the robustness of RT on general-purpose semantic understanding; 
(ii) Semantic reasoning subtasks from BRIGHT (e.g., Psy, TheoT), used to assess its gains on multi-step reasoning;
(iii) Recruitment matching tasks from PJBenchmark (JD2CV, JD2JD, etc.), used to examine its effectiveness in real-world semantic reasoning scenarios.

RT is applied only on the query $x_q$; documents $x_d$ is fixed. We set the number of iterative steps to $T\!\in\!\{1,2,\ldots,10\}$, where $T{=}1$ is the single-step baseline, and larger values correspond to deeper temporal unrolling. All experiments use the same hardware and batch settings to ensure fairness and comparability.

\subsection{Performance and Ablation Experiments}

Overall, RT remains stable on general-purpose semantic understanding tasks and achieves consistent gains on semantic reasoning tasks. On C-MTEB STS subtasks, RT is essentially on par with the baseline \ref{tab:cmteb-sts-results} , indicating that temporal unrolling does not degrade representation quality in standard similarity benchmarks. Across BRIGHT subtasks, RT improves the overall average from 22.9 to 23.1 ($\approx$ +1\%, Table \ref{tab:bright-results}), indicating that additional reasoning steps bring marginal benefits to complex retrieval. On PJBenchmark, RT achieves substantial gains in the Algorithm domain (e.g., JD2CV: +19\%, CV2CV: +21\%), while performance in the Finance domain slightly declines by 2\% \ref{tab:pjbenchmark-results}, suggesting that its benefits depend on task structure and domain noise.

\begin{table}[h]
  \centering
  \caption{Performance on C-MTEB STS subtasks}
  \label{tab:cmteb-sts-results}
  \small
  \resizebox{0.8\textwidth}{!}{%
    \begin{tabular}{lccccccc}
      \toprule
      \textbf{Model} & \textbf{STS} & \textbf{ATEC} & \textbf{BQ} & \textbf{LCQMC} & \textbf{PAWSX} & \textbf{QBQTC} & \textbf{STSB} \\
      \midrule
      Qwen3-Embedding-8B & \textbf{52.70} & \textbf{53.81} & \textbf{75.87} & \textbf{80.11} & \textbf{54.10} & 40.55 & \textbf{86.31} \\
      Qwen3-Embedding-8B + RT & 52.64 & 53.74 & 75.70 & 80.02 & 53.79 & \textbf{40.60} & 86.30 \\
      \bottomrule
    \end{tabular}%
  }
\end{table}

\begin{table}[h]
  \centering
  \caption{Performance on BRIGHT}
  \label{tab:bright-results}
  \resizebox{0.95\textwidth}{!}{%
    \begin{tabular}{lcccccccccccc|c}
      \toprule
      \multirow{2}{*}{\textbf{Model}} 
      & \multicolumn{7}{c}{\textbf{StackExchange}} 
      & \textbf{Code} 
      & \multicolumn{4}{c}{\textbf{Theorem-based}} 
      & \multirow{2}{*}{\textbf{Average}} \\
      \cmidrule(lr){2-8} \cmidrule(lr){9-9} \cmidrule(lr){10-13}
      & Bio & Earth & Econ & Psy & Rob & Stack & Sus 
      & Pony 
      & Leet & AoPS & TheoQ & TheoT 
      & \\
      \midrule
      Qwen3-Embedding-8B 
      & \textbf{21.3} & 32.0 & \textbf{19.4} & 27.7 & \textbf{15.8} & \textbf{18.8} & \textbf{18.3} 
      & 1.0 
      & \textbf{32.4} & 9.6 & 39.6 & 39.0 
      & 22.9 \\
      Qwen3-Embedding-8B + RT 
      & 21.1 & \textbf{33.9} & 19.1 & \textbf{28.1} & 15.7 & 18.0 & 18.1 
      & \textbf{1.1} 
      & 32.1 & \textbf{9.6} & \textbf{40.1} & \textbf{40.5} 
      & \textbf{23.1} \\
      \bottomrule
    \end{tabular}%
  }
\end{table}

\begin{table}[h]
  \centering
  \caption{Performance on PJBenchmark}
  \label{tab:pjbenchmark-results}
  \small
  \resizebox{0.6\textwidth}{!}{%
    \begin{tabular}{lccc|c}
      \toprule
      & \multicolumn{3}{c|}{\textbf{Algorithmic}} & \textbf{Finance} \\
      \cmidrule(lr){2-4} \cmidrule(lr){5-5}
      \textbf{Model} & \textbf{JD2JD} & \textbf{JD2CV} & \textbf{CV2CV} & \textbf{JD2CV} \\
      \midrule
      Qwen3-Embedding-8B & 60.98 & 62.77 & 52.14 & \textbf{62.66} \\
      Qwen3-Embedding-8B + RT & \textbf{62.07} & \textbf{74.33} & \textbf{63.26} & 61.30 \\
      \bottomrule
    \end{tabular}%
  }
\end{table}

\begin{figure}[H]
  \centering
  \includegraphics[width=0.5\textwidth]{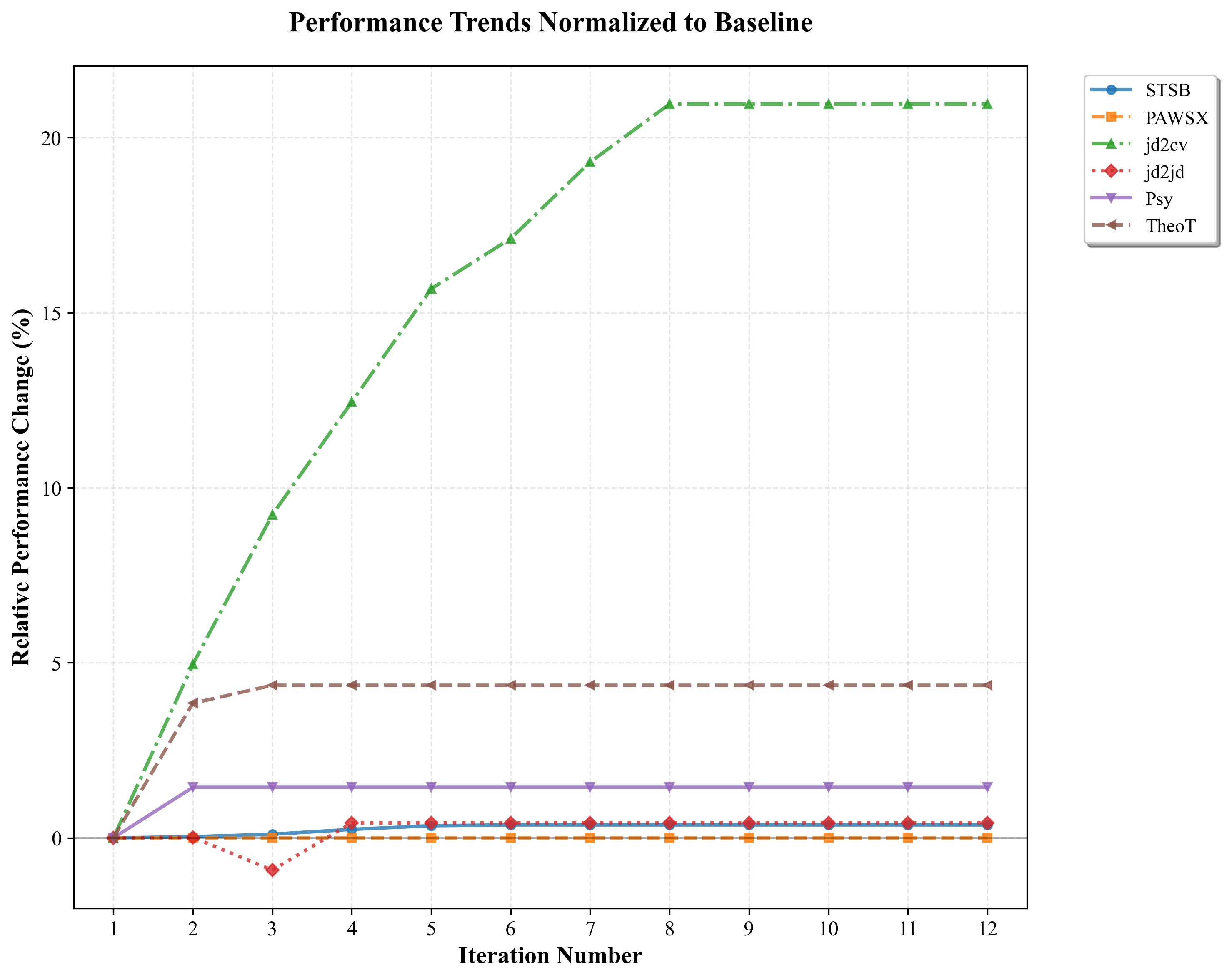}
  \caption{Performance of the RT Method across Tasks and Iteration Steps}
  \label{fig:rt_performance}
\end{figure}

We examine the effect of the iteration step number $T$ \ref{fig:rt_performance} on performance. Results indicate that the most notable gains occur at $T=2$--$3$, after which performance saturates for $T>3$, aligning with the notion of trading time steps for effective depth. On general semantic understanding tasks such as STS-B in C-MTEB, RT performs on par with the baseline, introducing no additional noise. In contrast, for high-complexity semantic reasoning tasks such as BRIGHT and PJBenchmark, RT brings substantial improvements, suggesting that temporal unrolling is particularly beneficial in scenarios requiring multi-step or compositional reasoning.

We further compare encoder-only and decoder-only architectures under the same RT setup on the PJBenchmark JD2CV task (Table~\ref{tab:pjbenchmark-jd2cv}). Qwen3-Embedding-8B improves from 62.77 to 74.33, whereas bge-large-zh-v1.5 rises only modestly from 21.66 to 23.06. These results suggest that RT’s temporal unrolling mechanism aligns more naturally with autoregressive (decoder-only) architectures. In contrast, encoder-only models can simulate iteration via an outer loop, but their architectural constraints and state-update mechanisms limit performance gains. This supports the hypothesis that recursion and variable-length computation are key to complex reasoning, and that decoder-only embedding models possess a natural advantage for temporal unrolling.

\begin{table}[h]
  \centering
  \caption{PJBenchmark JD2CV task comparison}
  \label{tab:pjbenchmark-jd2cv}
  \small
  \resizebox{0.35\textwidth}{!}{%
    \begin{tabular}{lc}
      \toprule
      \textbf{Model} & \textbf{JD2CV} \\
      \midrule
      Qwen3-Embedding-8B & 62.77 \\
      Qwen3-Embedding-8B + RT & \textbf{74.33} \\
      bge-large-zh-v1.5 & 21.66 \\
      bge-large-zh-v1.5 + RT & 23.06 \\
      \bottomrule
    \end{tabular}
  }
\end{table}

\subsection{Analysis and Discussion}

The RT method achieves strong improvements on semantic reasoning tasks while remaining comparable on general-purpose semantic understanding benchmarks. The gains peak at $T=2$--$3$ and level off for larger $T$, supporting the idea that additional time steps increase effective reasoning depth. Decoder-only models benefit more than encoder-only ones, showing that autoregressive architectures more effectively integrate previously generated reasoning trajectories. The improvement scales with task complexity, nearly zero for STS-B but substantial for BRIGHT and PJBenchmark, showing that temporal unrolling helps the model compose reasoning more effectively in hidden space.

Although the RT method performs well on complex tasks, it also has notable limitations. First, RT is an inference-only extension and lacks explicit credit assignment \cite{setlur2024rewarding}. As a result, the model cannot assess the effectiveness of each computation step, and the additional iterations only make limited adjustments within the existing parameter space, without forming new strategies or long-term memory. Second, RT shows little improvement on shallow or noisy tasks, and its reasoning trajectories tend to drift on long or loosely organized inputs, mainly due to the absence of explicit structural constraints and memory management \cite{sastre2025memorytokens}. Finally, RT is sensitive to both the number of iterations and the state aggregation scheme: too many steps increase latency, while too few reduce reasoning depth. Adaptive step scheduling and early-stopping mechanisms are therefore needed in practice to maintain stability and efficiency.

\section{Conclusion and Future Work}
\label{sec:conclusion}

We propose RT, a test-time method that enhances the reasoning capability of text representation models, addressing their limited computational depth in semantic reasoning tasks. Experiments show that RT significantly improves performance on semantic reasoning benchmarks such as BRIGHT and PJBenchmark, while maintaining stable results on general-purpose semantic understanding tasks such as C-MTEB, confirming the effectiveness of temporal unrolling for complex semantic reasoning. RT treats semantic reasoning as a temporally unfolded computational process, enabling the model to acquire additional reasoning steps at test time and thereby extend its effective reasoning depth without modifying parameters. Our results confirm that recursion and variable-length computation underlie complex reasoning, and that decoder-only architectures are inherently more effective for semantic reasoning. Although RT method performs well across various tasks, its still relies on heuristic control without fine-grained learning or validation. Future directions include integrating lightweight retraining or reinforcement learning \cite{dong2025rpt} for reward modeling \cite{lightman2024letsverify} and developing more efficient temporal unrolling strategies under decoder-only architectures.

Prior research indicates that large reasoning models acquire their reasoning ability through exposure to reasoning trajectories during pre-training \cite{ruis2025procedural,hatamizadeh2025rlp}. This insight likely explains RT’s effectiveness on decoder-only models such as Qwen3-Embedding-8B. At present, our experiments focus mainly on semantic reasoning tasks, without extending to symbolic or arithmetic reasoning, and the range of evaluated models remains limited. Therefore, further studies are required to disentangle whether the observed improvements arise from differences in architecture, task type, or training data, and to verify the general advantage of decoder-only architectures under the RT framework.

\bibliography{references}

\appendix

\end{document}